\documentclass[twocolumn,10pt]{asme2ej}

\usepackage{epsfig} 
\usepackage{epstopdf}
\usepackage{cite}
\usepackage[cmex10]{amsmath}
\usepackage{amssymb}
\usepackage{algorithm}
\usepackage{algpseudocode}
\usepackage{array}
\usepackage{tikz}
\newcommand*\circled[1]{\tikz[baseline=(char.base)]{
            \node[shape=circle,draw,inner sep=2pt] (char) {#1};}}
\title{A Sequential Two-Step Algorithm for Fast Generation of Vehicle Racing Trajectories}

\author{Nitin R. Kapania \thanks{Address all correspondence to this author.}
    \affiliation{
	Dept. of Mechanical Engineering\\
	Stanford University\\
	Stanford, CA 94305\\
	nkapania@stanford.edu
    }	
}

\author{John Subosits
    \affiliation{Dept. of Mechanical Engineering\\
	Stanford University\\
	Stanford, CA 94305\\
	subosits@stanford.edu
    }
}

\author{J. Christian Gerdes
    \affiliation{Dept. of Mechanical Engineering\\
	Stanford University\\
	Stanford, CA 94305\\
	gerdes@stanford.edu
    }
}

\begin{document}

\maketitle

\begin{abstract}
{\it The problem of maneuvering a vehicle through a race course in minimum time requires computation of 
 both longitudinal (brake and throttle) and lateral (steering wheel) control inputs.    
 Unfortunately, solving the resulting nonlinear optimal control problem is typically computationally expensive and infeasible for real-time trajectory planning.
 This paper presents an iterative algorithm that divides the path generation
 task into two sequential subproblems that are significantly easier to solve. Given an initial path through the race track, the algorithm
 runs a forward-backward integration scheme to determine the minimum-time longitudinal speed profile, subject to
 tire friction constraints. With this fixed speed profile, the algorithm updates the vehicle's path by solving a convex optimization problem 
 that minimizes the resulting path curvature while staying within track boundaries and obeying affine, time-varying vehicle dynamics constraints.
 This two-step process is repeated iteratively until the
 predicted lap time no longer improves. While providing no guarantees of convergence or a globally optimal solution, 
 the approach performs very well when validated on the Thunderhill Raceway
 course in Willows, CA. The predicted lap time converges after four to five iterations, with each iteration over the full 4.5 km race course requiring
 only thirty seconds of computation time on a laptop computer. The resulting trajectory is experimentally driven at the race circuit with
 an autonomous Audi TTS test vehicle, and the resulting lap time and racing line is comparable to both a nonlinear gradient
 descent solution and a trajectory recorded from a professional racecar driver. The experimental results indicate that the proposed method is a viable option for 
 online trajectory planning in the near future.}
\end{abstract}

\section{Introduction}
The problem of calculating the minimum lap time trajectory for a given vehicle and race track has been studied over the last several decades
in the control, optimization, and vehicle dynamics communities.  Early research by Hendrikx et al. \cite{hendrikx} in 1996 used 
 Pontryagin's minimum principle to derive coupled differential equations to solve for the minimum-time trajectory
of a vehicle lane change maneuver. The minimum lap time problem drew significant interest from professional racing teams, and
Casanova \cite{casanova} published a method in 2000 capable of simultaneously optimizing both the path and speed profile
for a fully nonlinear vehicle model using nonlinear programming (NLP). Kelly \cite{kelly} further extended the results from Casanova by considering
the physical effect of tire thermodynamics and applying more robust NLP solution methods such as Feasible Sequential Quadratic Programming. More recently,
Perantoni and Limebeer \cite{perantoni} showed that the computational expense could be significantly reduced by 
applying curvilinear track coordinates, non-stiff vehicle dynamics, and the use of smooth computer-generated
analytic derivatives. The authors simulated the optimal vehicle lap on the Catalunya race circuit with a spatial discretization
of two meters and a solve time of around 15 minutes. 

More recently, the development of autonomous vehicle technology at the industry and academic level has led to research on optimal path
 planning algorithms that can be used for driverless cars. Theodosis and Gerdes published a gradient descent approach for determining time-optimal
  racing lines, with the racing line constrained to be composed of a fixed number of clothoid segments \cite{theodosis}. Given the computational expense
 of performing nonlinear optimization, there has also been a significant research effort to find approximate methods that provide fast lap times.
 Timings and Cole \cite{timings} formulated the minimum lap time problem into a model predictive control (MPC) problem by linearizing the nonlinear vehicle
 dynamics at every time step and approximating the minimum-time objective by maximizing distance traveled along the path centerline. The resulting racing line
 for a 90 degree turn was simulated next to an NLP solution. Gerdts et al.  \cite{gerdts} proposed a similar receding horizon approach, where distance along a reference path was maximized over a series of locally optimal optimization
 problems that were combined with continuity boundary conditions. One potential drawback of the model predictive control approach is that an optimization
 problem must be reformulated and solved at every time step, which can still be computationally expensive. For example, Timings and Cole reported a computation time of 900 milliseconds
 per 20 millisecond simulation step with the CPLEX quadratic program solver on a desktop PC.

 While experimental validation was reported only by \cite{theodosis} and \cite{gerdts}, all of the aforementioned methods are 
 feasible for experimental implementation, as an autonomous vehicle can apply a closed-loop controller to follow a time-optimal vehicle trajectory computed offline. 
 However, there are  significant benefits to developing a fast trajectory generation algorithm that can approximate the globally optimal trajectory in real-time. 
 If the algorithm runtime is small compared to the actual lap time, the algorithm can run as a real-time trajectory planner and find a fast racing line 
 for the next several turns of the racing circuit. This would allow the trajectory planner to modify the desired path based on the motion of competing race vehicles and 
 estimates of road friction, tire wear, engine/brake dynamics and other parameters learned over several laps of racing. Additionally, the fast trajectory algorithm can
 be used to provide a very good initial trajectory for a nonlinear optimization method.  
  
 This paper therefore presents an experimentally validated iterative algorithm that generates vehicle racing trajectories with low computational expense. 
 To decrease computation time, the combined lateral/longitudinal optimal control problem is replaced by two sequential sub-problems 
 where minimum-time longitudinal speed inputs are computed given a fixed vehicle path, and then the vehicle path is updated given the fixed speed commands. 
 
 The following section presents a mathematical framework for the trajectory generation
 problem and provides a linearized five-state model for the planar dynamics of a racecar following a set of speed and steering inputs, with lateral and heading error states computed with respect to a fixed path.
 Section \ref{sec:VP} describes the method of finding the minimum-time speed inputs given a fixed path.  While this task has been recently formulated as a convex optimization problem \cite{lipp}, a forward-backward
 integration scheme based on prior work \cite{subosits} is used instead.  Section \ref{sec:UPDATE} describes a method for updating the racing path given the fixed speed inputs using convex optimization, where the curvature norm
 of the driven path is explicitly minimized. The complete algorithm is outlined in Section \ref{sec:IMPLEMENT}, and the racing trajectory generated on the Thunderhill Raceway circuit
 in Willows, CA is compared to results from both a nonlinear optimization and data recorded from a professional racecar driver. In Section \ref{sec:EXP}, the racing trajectory is then tested experimentally in
 an autonomous Audi TTS testbed via a previously published closed-loop path following controller \cite{kapaniaVSD}. The resulting lap time compares well with the lap time
 recorded for the nonlinear optimization trajectory. Section \ref{sec:DISCUSSION} concludes by discussing 
 future implementation of the algorithm in a real-time path planner.  

\section{Path Description and Vehicle Model}
\label{sec:PATH}
\begin{figure}
\centering
\includegraphics[width=3.4in]{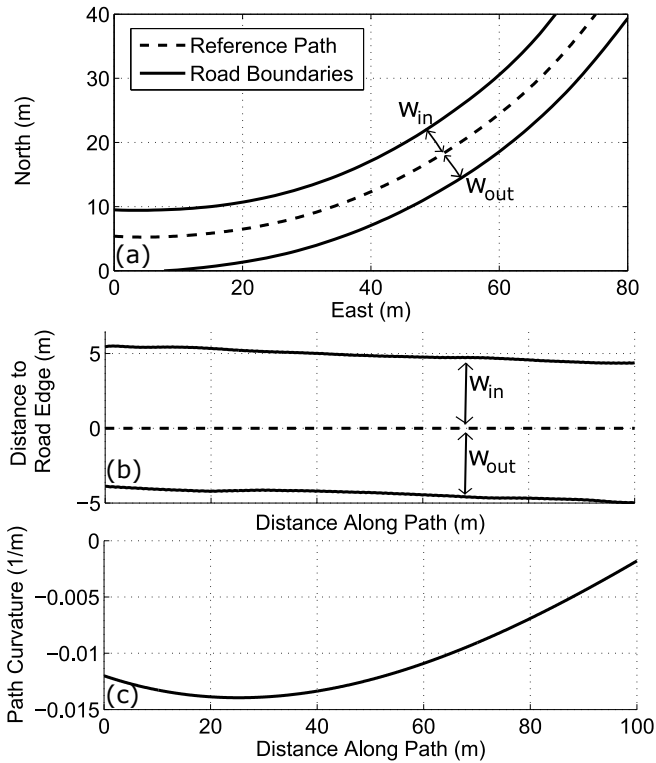}
\caption{(a) View of a sample reference path and road boundaries, plotted in the East-North Cartesian frame (b) Lateral distance from path to inside road edge (positive) and outside road edge (negative) as a function
of distance along path. (c) Curvature as a function of distance along path.}
\label{fig:worldInfo}
\end{figure}

Figure~\ref{fig:worldInfo} describes the parameterization of the reference path that the vehicle will follow. The reference path is most intuitively described in Fig.~\ref{fig:worldInfo}(a) as a 
smooth curve of Cartesian East-North coordinates, with road boundaries represented by similar Cartesian curves. However, for the purposes of quickly generating
a racing trajectory, it is more convenient to parameterize the reference path as a curvature profile $K$ that is a function of distance along the path $s$ (Fig.~\ref{fig:worldInfo}(c)). Additionally, it is 
convenient to store the road boundary information as two functions $w_\mathrm{in}(s)$ and $w_\mathrm{out}(s)$, which correspond to the lateral distance from the path at $s$
 to the inside and outside road boundaries, respectively (Fig.~\ref{fig:worldInfo}(b)). This maximum lateral distance representation will be useful when constraining the generated racing path to lie within the road 
 boundaries. The transformation from the local $s$, $K$ coordinate frame to 
the global Cartesian coordinates $E$, $N$ are given by the Fresnel integrals:
\begin{subequations}
\label{eq:fresnel}
\begin{align}
	E(s) &= \int_0^s  -\sin(\Psi_\mathrm{r}(z)) dz \\
	N(s) &= \int_0^s   \cos(\Psi_\mathrm{r}(z)) dz \\
	\Psi_\mathrm{r}(s) &= \int_0^s K(z) dz \label{eq:balls}
\end{align}
\end{subequations}
where $\Psi_\mathrm{r}(s)$ is the heading angle of the reference path and $z$ is a dummy variable. 

\begin{figure}
\centering
\includegraphics[width=2.5in]{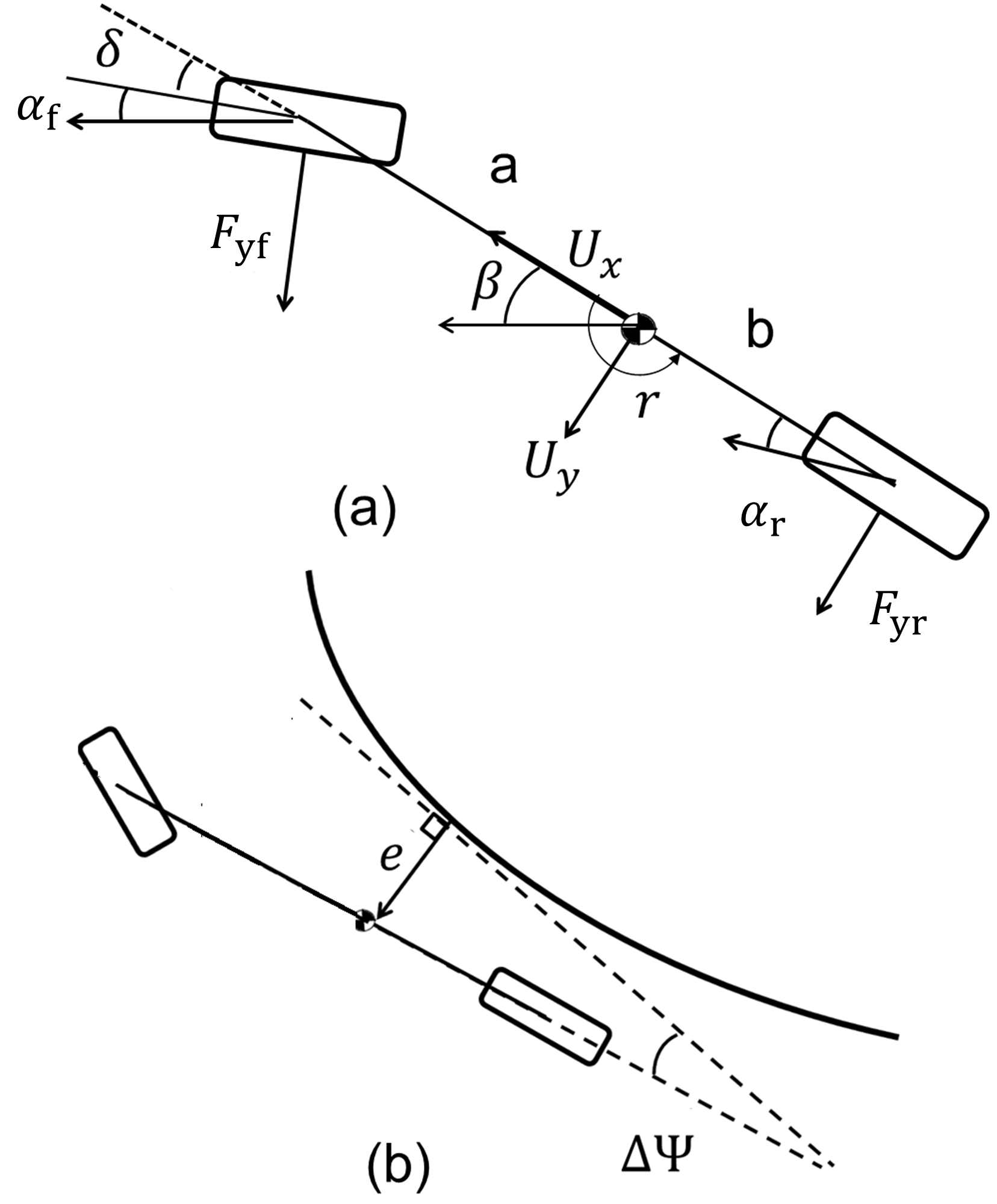}
\caption{(a) Schematic of bicycle model. (b) Diagram showing lateral path deviation $e$ and path heading error $\Delta\Psi$ states.}
\label{fig:bikemodel}
\end{figure}

With the reference path defined in terms of $s$ and $K$, the 
next step is to define the dynamic model of the vehicle. For the purposes of trajectory generation, we assume the vehicle dynamics are given 
by the planar bicycle model in Fig.~\ref{fig:bikemodel}(a), with yaw rate $r$ and sideslip $\beta$ states describing the lateral dynamics. Additionally, the vehicle's offset from
the reference path is given by the path lateral deviation state $e$ and path heading error state $\Delta\Psi$ (Fig.~\ref{fig:bikemodel}(b)). Linearized equations of
motion for all four states are given by:
\begin{subequations}
\label{eq:bm}
\begin{align}
	\dot{\beta} &= \frac{F_\mathrm{yf}+F_\mathrm{yr}}{mU_x} - r \qquad \dot{r} = \frac{aF_\mathrm{yf} - bF_\mathrm{yr}}{I_z} \label{bm1} \\
	\dot{e} &= U_x (\beta + \Delta\Psi) \qquad \Delta\dot{\Psi} = r - U_xK \label{eq:bm2} 
\end{align}
\end{subequations}
where $U_x$ is the vehicle forward velocity and $F_\mathrm{yf}$ and $F_\mathrm{yr}$ are the front and rear lateral tire forces. 
The vehicle mass and yaw inertia are denoted by $m$ and $I_z$, and the geometric parameters $a$ and $b$ are shown in Fig.~\ref{fig:bikemodel}(a). Note
that while the vehicle longitudinal dynamics are not explicitly modeled, the bicycle model does allow for time-varying values of $U_x$.

\section{Velocity Profile Generation Given Fixed Reference Path}
\label{sec:VP}

\begin{figure}
\centering
\includegraphics[width=3.3in]{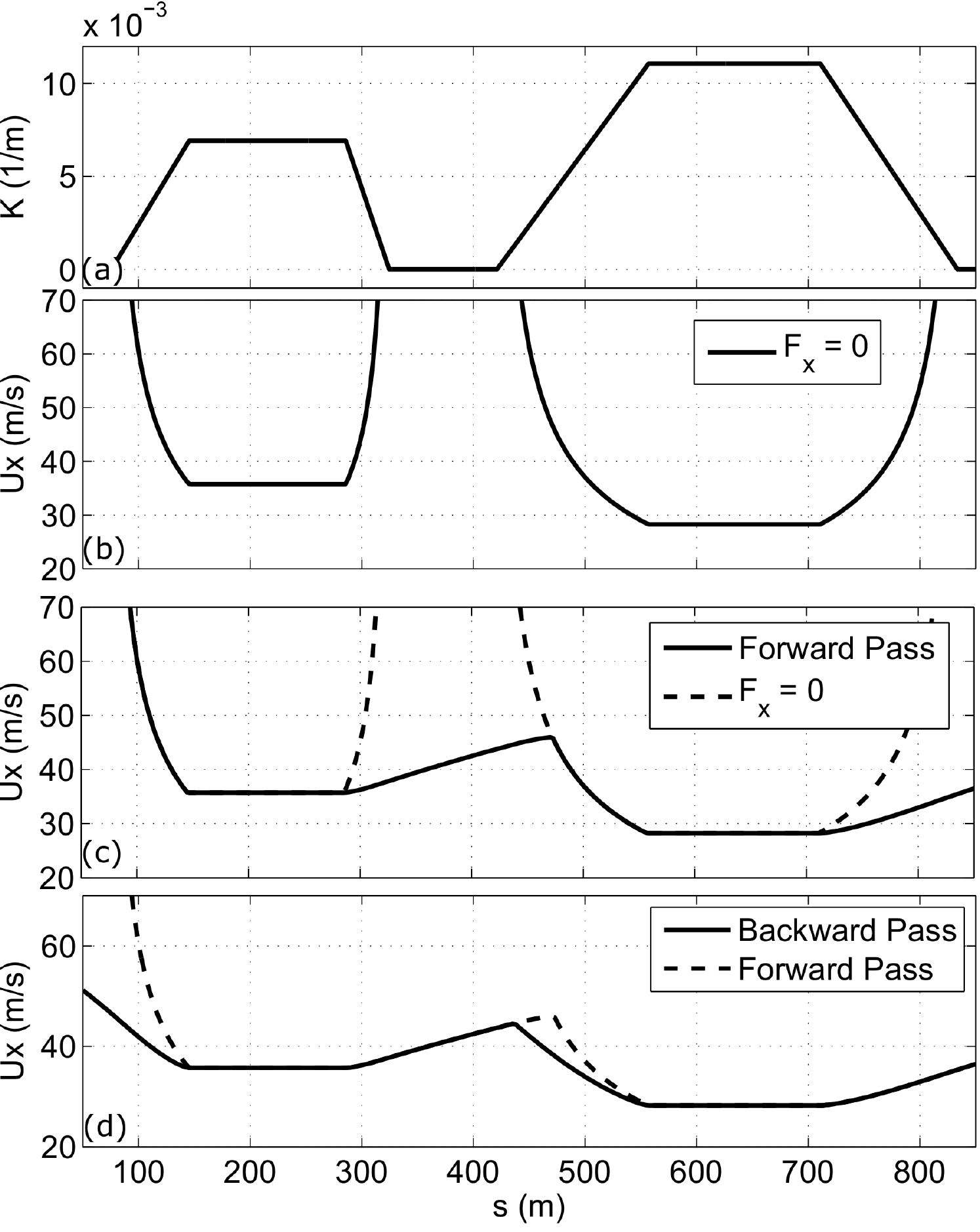}
\caption{(a) Sample curvature profile. (b) Velocity profile given zero longitudinal force. (c) Velocity profile after forward pass (d) Final velocity profile after backward pass. }
\label{fig:VPgen}
\end{figure}

Given a fixed reference path described by $s$ and $K$, the first algorithm step is to find the minimum-time speed profile the vehicle can achieve
without exceeding the available tire-road friction. The approach taken in this paper is a ``three-pass" approach described in complete detail by 
Subosits and Gerdes \cite{subosits}, and originally inspired by work from Velenis and Tsiotras \cite{velenis}. 
Given the lumped front and rear tires from Fig.~\ref{fig:bikemodel}(a), the available longitudinal
force $F_\mathrm{x}$ and lateral force $F_\mathrm{y}$ at each wheel is constrained by the friction circle: 
\begin{subequations}
\label{eq:tireforce}
\begin{align}
	F^2_\mathrm{xf} + F_\mathrm{yf}^2 &\leq (\mu F_\mathrm{zf})^2\\
	F^2_\mathrm{xr} + F_\mathrm{yr}^2 &\leq (\mu F_\mathrm{zr})^2
\end{align}
\end{subequations}
where $\mu$ is the tire-road friction coefficient and $F_\mathrm{z}$ is the available normal force. The first pass of the speed profile generation aims to find the maximum permissible steady 
state vehicle speed given zero longitudinal force. For the simplified case where weight transfer and topography effects are neglected, this is given by:
\begin{equation}
\label{eq:steadystate}
	U_x(s) = \sqrt{\frac{\mu g}{|K(s)|}}
\end{equation}
where the result in (\ref{eq:steadystate}) is  obtained by setting $F_\mathrm{yf} = \frac{mb}{a+b}U_x^2K$ and $F_\mathrm{zf} = \frac{mgb}{a+b}$.
 The results of this first pass for the sample curvature profile in Fig.~\ref{fig:VPgen}(a) are shown in Fig.~\ref{fig:VPgen}(b).
 The next step is a forward 
integration step, where the velocity of a given point is determined by the velocity of the previous point and the available longitudinal force $F_{x,\mathrm{max}}$ for acceleration.
This available longitudinal force is calculated in \cite{subosits} by accounting for the vehicle engine force limit and the lateral force demand on all tires due to the road curvature:
\begin{equation}
\label{eq:forwardint}
	U_x(s+\Delta s) =\sqrt{U^2_x(s)+\mathrm{2}\frac{F_{x,\mathrm{accel,max}}}{m}\Delta s}
\end{equation}
A key point of the forward integration step is that at every point, the value of $U_x(s)$ is compared to the corresponding value from (\ref{eq:steadystate}), and
the minimum value is taken. The result is shown graphically in Fig.~\ref{fig:VPgen}(c). Finally, the backward integration step occurs, where the available
longitudinal force for deceleration is again constrained by the lateral force demand on all tires:
\begin{equation}
\label{eq:backwardsint}
	U_x(s-\Delta s) = \sqrt{U^2_x(s)-\mathrm{2}\frac{F_{x,\mathrm{decel,max}}}{m}\Delta s}
\end{equation}
The value of $U_x(s)$ is then compared to the corresponding value from (\ref{eq:forwardint}) for each point along the path, and the minimum
value is chosen, resulting in the final velocity profile shown by the solid line in Fig.~\ref{fig:VPgen}(d). While treatment of three-dimensional topography 
effects are not described in this paper, the method described in \cite{subosits} and
used for the experimental data collection determines the normal and lateral tire forces $F_\mathrm{z}$ and $F_\mathrm{y}$ at each point along the path 
by accounting for weight transfer, road bank and grade. 

\section{Updating Path Given Fixed Velocity Profile}
\label{sec:UPDATE}
\subsection{Overall Approach and Minimum Curvature Heuristic}
The second step of the trajectory generation algorithm takes the original reference path $K(s)$ and corresponding velocity profile $U_x(s)$
as inputs, and modifies the reference path to obtain a new, ideally faster, racing line. Sharp \cite{sharp} suggests a general approach for modifying 
an initial path to obtain a faster lap time by taking the original path and velocity profile and incrementing the speed uniformly
by a small, constant ``learning rate." An optimization problem is then solved to find a new reference path and control inputs that allow the vehicle to
drive at the higher speeds without driving off the road. If a crash is detected, the speed inputs are locally reduced around the crash site and the process is repeated.

However, one challenge with this approach is that it can take several hundred iterations of locally modifying the vehicle speed profile, detecting crashes, and 
 modifying the reference path to converge to a fast lap time. An alternative approach is to modify the reference path in one step by solving a single 
 optimization problem. The lap time $t$ for a given racing line is provided by the following equation: 
 
 \begin{equation}
t = \int_0^L\frac{ds}{U_x(s)}
\label{integrateEq}
\end{equation}
 
 Equation (\ref{integrateEq}) implies that minimizing the vehicle lap time requires simultaneously minimizing the total path length $L$ while maximizing
 the vehicle's longitudinal velocity $U_x$. These are typically competing objectives, as lower curvature (i.e. higher radius) paths can result in longer path lengths but higher
 vehicle speeds when the lateral force capability of the tires is reached, as shown in (\ref{eq:steadystate}). As mentioned in the introduction, time-intensive nonlinear programming
 is required to manage this trade-off and directly minimize (\ref{integrateEq}). 
 
 The proposed approach is therefore to simplify the cost function by only minimizing the norm of the vehicle's driven curvature
 $K(s)$ at each path modification step. Path curvature can be easily formulated as a convex function with respect to the vehicle state vector $x$, 
 enabling the path modification step to be easily solved by leveraging the computational speed of convex optimization. 
 
 However, minimizing curvature is not the same as minimizing lap time and provides
 no guarantee of finding the time-optimal solution. The proposed cost function relies on the hypothesis that a path with minimum curvature is a good approximation for
 the minimum-time racing line. Lowering the curvature of the racing line is more important than minimizing
 path length for most race courses, as the relatively narrow track width provides limited room to shorten the overall path length. Simulated and experimental results
 in Sections \ref{sec:IMPLEMENT} and \ref{sec:EXP} will validate this hypothesis by showing similar lap time performance when compared to
 a gradient descent method that directly minimizes lap time. One reason for this comparable performance is that nonlinear methods are
 typically sensitive to initial conditions and are themselves only guaranteed to find a locally optimal solution.  

\subsection{Convex Problem Formulation}

\begin{figure}
\centering
\includegraphics[width=3.7in]{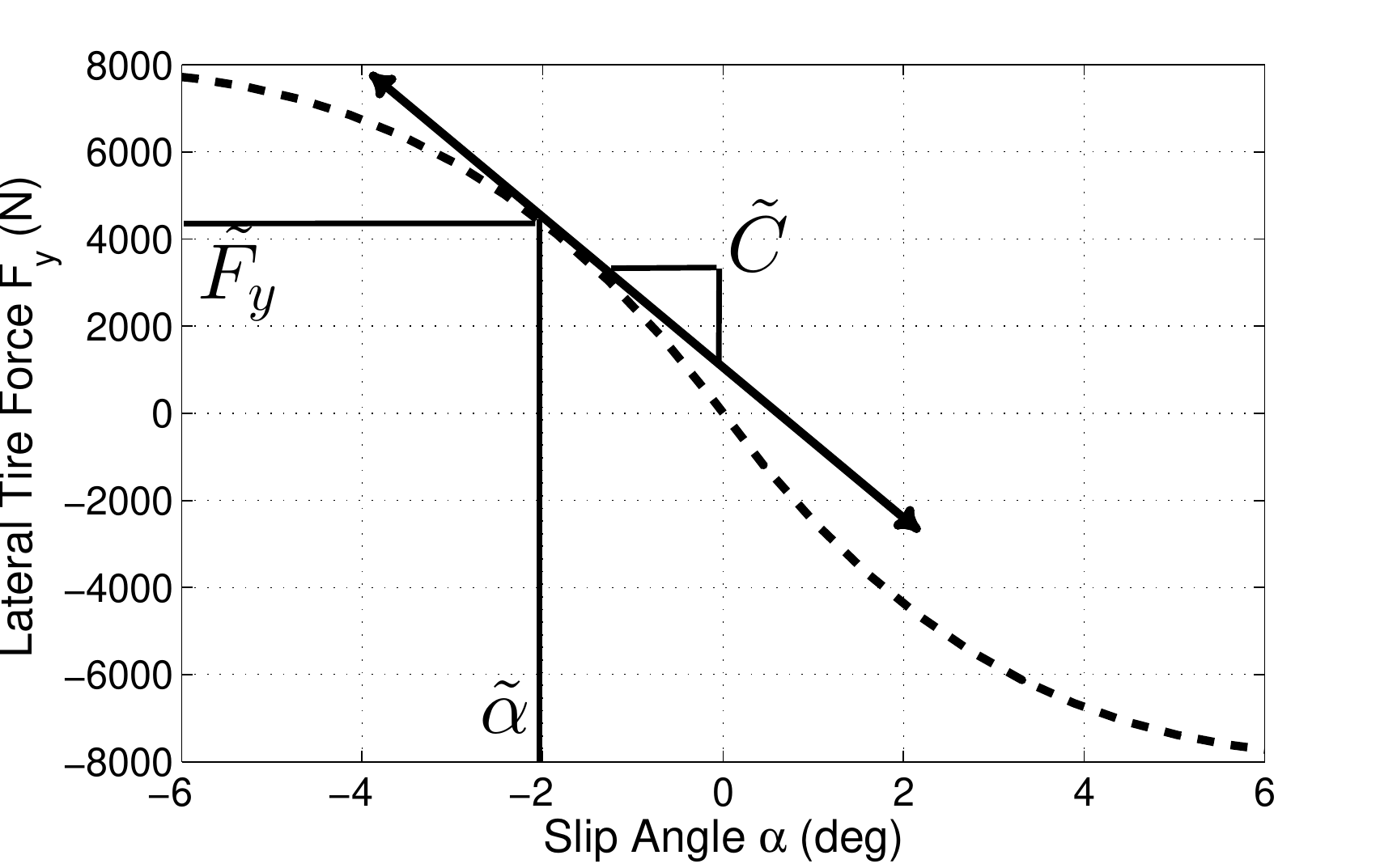}
\caption{Nonlinear tire force curve given by Fiala model, along with affine tire model linearized at $\alpha = \tilde{\alpha}$. }
\label{fig:fiala}
\end{figure}

Formulating the path update step as a convex optimization problem requires an affine, discretized form of the
bicycle model in (\ref{eq:bm}). The equations of motion are nonlinear because
 the front and rear lateral tire forces become saturated as the vehicle drives near the limits of tire adhesion. 
The well-known brush tire model \cite{Pacejka2012} captures the force saturation of tires as a function of lateral tire slip angle $\alpha$ as follows:
\small
\begin{eqnarray}
\label{eq:fiala}
	F_\mathrm{y\diamond}&=&\begin{cases} -C_{\diamond}\tan\alpha_\diamond + \frac{C_\diamond^2}{3\mu F_\mathrm{z\diamond}} |\tan\alpha_\diamond| \tan\alpha_\diamond \\ \hspace{10mm}- \frac{C_\diamond^3}{27\mu^2F_\mathrm{z\diamond}^2}\tan^3\alpha_\diamond,
\hspace{8mm}  |\alpha_\diamond| < \arctan{\left(\frac{3\mu F_\mathrm{z\diamond}}{C_\diamond}\right)} \\ \\ -\mu F_\mathrm{z\diamond}\text{sgn} \ \alpha_\diamond, \hspace{36mm} \mathrm{otherwise} \end{cases}
\end{eqnarray}
\normalsize
where the symbol $\diamond \in [\mathrm{f},\mathrm{r}]$ denotes the lumped front or rear tire, and $C_\diamond$ is the corresponding tire cornering stiffness. 
The linearized tire slip angles $\alpha_\mathrm{f}$ and $\alpha_\mathrm{r}$ are functions of the vehicle lateral states and the steer angle
input, $\delta$:
\begin{subequations}
\begin{align}
	\alpha_\mathrm{f} &= \beta + \frac{ar}{U_x} - \delta\\
	\alpha_\mathrm{r} &= \beta - \frac{br}{U_x}
\end{align}
\end{subequations}
The tire model (\ref{eq:fiala}) can be linearized at every point along the reference path assuming steady state cornering conditions:
\begin{subequations}
\begin{align}
	F_\mathrm{y\diamond} &= \tilde{F}_\mathrm{y\diamond} - \tilde{C}_\diamond(\alpha_\diamond - \tilde{\alpha}_\diamond) \\
	\tilde{F}_\mathrm{y\diamond} &= \frac{F_\mathrm{z\diamond}}{g} U_x^2K
\label{eqn:ftil}
\end{align}
\end{subequations}
with parameters $\tilde{F}_\mathrm{y}$, $\tilde{\alpha}$ and $\tilde{C}$ shown in Fig.~\ref{fig:fiala}. The affine, continuous bicycle model with
steering input $\delta$ is then written in state-space form as:
\begin{subequations}
\label{eq:cts}
\begin{align}
	\dot{x}(t) &= A(t) x + B(t)\delta + d(t)\\
	 x &= [e \hspace{2mm} \Delta\Psi \hspace{2mm} r \hspace{2mm} \beta \hspace{2mm} \Psi]^T
\end{align}
\end{subequations}
where we have added a fifth state, vehicle heading angle $\Psi$, defined as the time integral of yaw rate $r$.
 This makes explicit computation of the minimum curvature path simpler. The state matrices $A(t)$, $B(t)$, and affine term $d(t)$ 
 are given by:
\begin{multline}
\label{eqn:Amatrix}
A(t)  =  \\
\left[\begin{matrix}
  0 & U_x(t) & 0 & U_x(t) & 0\\ 
  0 & 0 & 1 & 0 & 0 \\ 
  0 & 0  & \frac{-(a^2\tilde{C}_\mathrm{f}(t)+b^2\tilde{C}_\mathrm{r}(t))}{U_x(t)I_\mathrm{z}} & \frac{b\tilde{C}_\mathrm{r}(t) - a\tilde{C}_\mathrm{f}(t)}{I_\mathrm{z}} & 0 \\
  0 & 0  & \frac{b\tilde{C}_\mathrm{r}(t)-a\tilde{C}_\mathrm{f}(t)}{mU_x^2(t)}-1 & \frac{-(\tilde{C}_\mathrm{f}(t) + \tilde{C}_\mathrm{r}(t))}{mU_x(t)} & 0 \\
  0 & 0 & 1 & 0 & 0
 \end{matrix}\right]
 \end{multline}
\begin{align}
B(t) =\left[0 \hspace{2 mm} 0 \hspace{3 mm} \frac{a \tilde{C}_\mathrm{f}(t)}{I_\mathrm{z}} \hspace{3 mm}  \frac{\tilde{C}_\mathrm{f}(t)}{mU_x(t)} \hspace{3mm} 0\right]^T
\end {align}
\begin{align}
\label{eqn:Dmatrix}
d(t) = \left[\begin{matrix} 0 \\
               -K(t) U_x(t) \\ 
			    \frac{a\tilde{C}_\mathrm{f}(t)\tilde{\alpha}_\mathrm{f}(t) - b\tilde{C}_\mathrm{r}(t)\tilde{\alpha}_\mathrm{r}(t) + a\tilde{F}_\mathrm{yf}(t) - b\tilde{F}_\mathrm{yr}(t)}{I_z} \\
				\frac{\tilde{C}_\mathrm{f}(t)\tilde{\alpha}_\mathrm{f}(t) + \tilde{C}_\mathrm{r}(t)\tilde{\alpha}_\mathrm{r}(t) + \tilde{F}_\mathrm{yf}(t) + \tilde{F}_\mathrm{yr}(t)}{mU_x(t)}\\
				0
				\end{matrix}\right]
\end{align}

With the nonlinear model now approximated as an affine, time-varying model, updating the path is accomplished by solving the following
 convex optimization problem:

\begin{subequations}
\label{eq:OPT}
\begin{alignat}{3}
\underset{\delta, \hspace{.5mm} x}{\text{minimize}} \quad & \sum_{k} \left(\frac{\Psi_k - \Psi_{k-1}}{s_k - s_{k-1}}\right)^2 + \lambda(\delta_k - \delta_{k-1})^2 & \quad  \label{eq:obj}\\
{\text{subject to}} \quad & \rlap{$ x_{k+1}= A_k x_k + B_k \delta_k + d_k$} \label{jenny}\\
& \rlap{$w^\mathrm{out}_k \leq e_k \leq w^\mathrm{in}_k $} \label{bobby}\\
& \rlap{$x_1 = x_T$\label{keylani}}
\end{alignat}
\end{subequations}
where $k = 1 \dots T$ is the discretized time index, and $A_k$, $B_k$, and $d_k$ are discretized versions of the continuous state-space
equations in (\ref{eq:cts}). The objective function (\ref{eq:obj}) minimizes the curvature norm of the path driven by the vehicle, as path curvature is
the derivative of the vehicle heading angle with respect to distance along the path $s$ (\ref{eq:balls}). To maintain convexity of the objective
function, the term ${s_k - s_{k-1}}$ is a constant rather than a variable, and is updated for the next iteration after the optimization has been completed (see Section \ref{sec:IMPLEMENT}). 
Additionally, there is a regularization term with weight $\lambda$ added in the cost function to ensure a smooth steering profile for experimental 
implementation. 

\begin{figure}
\centering
\includegraphics[width=3.25in]{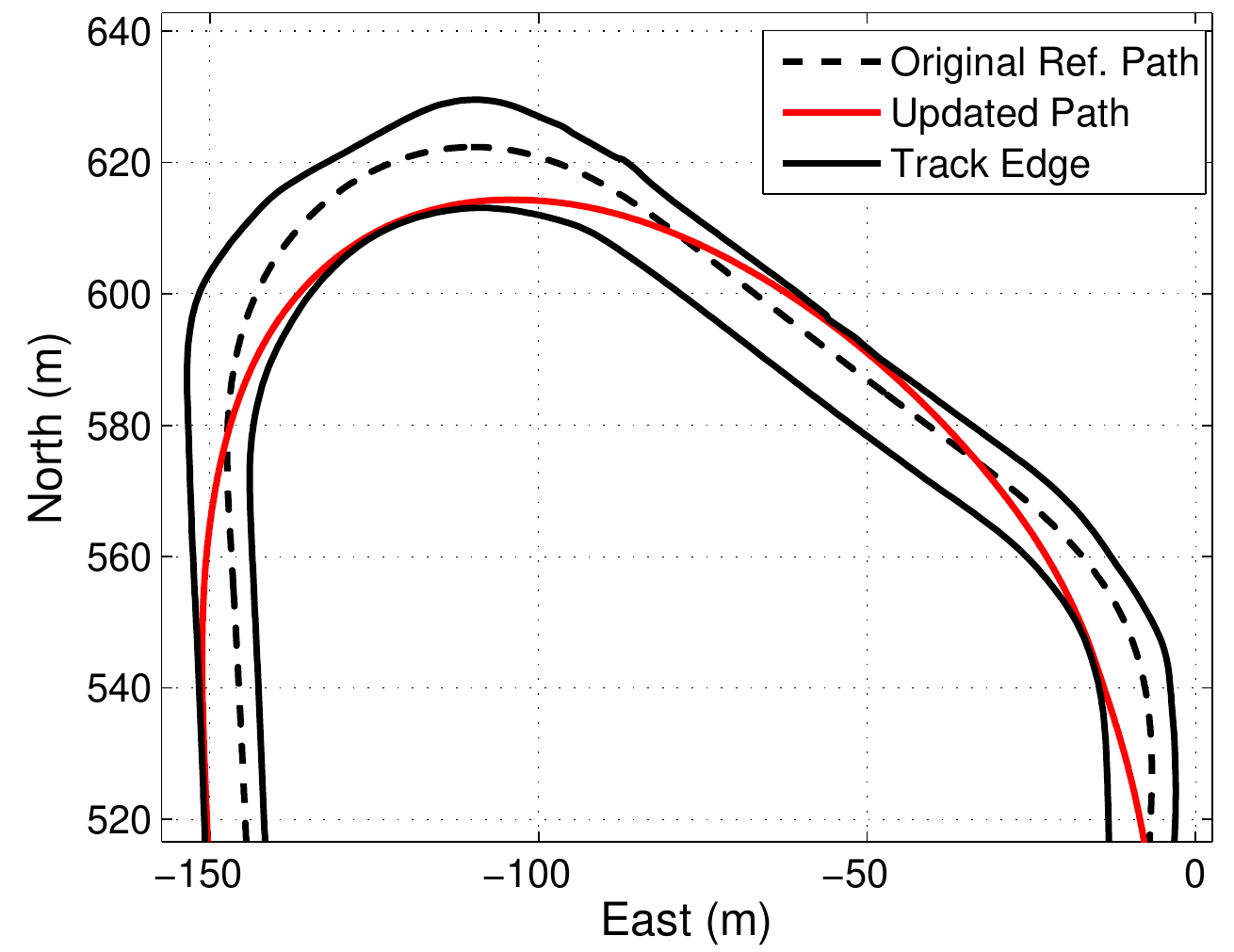}
\caption{Path update for an example turn.}
\label{fig:hairpin}
\end{figure}

The equality constraint (\ref{jenny}) ensures the vehicle follows the affine lateral dynamics. The inequality
 constraint (\ref{bobby}) allows the vehicle to deviate laterally from the reference path to find a new path with lower curvature, but
 only up to the road edges. Finally, the equality constraint (\ref{keylani}) is required for complete racing circuits to ensure the generated
 racing line is a continuous loop. The results of running the optimization 
 are shown for an example turn in Fig.~\ref{fig:hairpin}. The reference path starts out at the road centerline, and the optimization finds 
 a modified path that uses all the available width of the road to lower the path curvature.

\section{Algorithm Implementation and Simulated Results}
\label{sec:IMPLEMENT}
\subsection{Algorithm Implementation}
The final algorithm for iteratively
generating a vehicle racing trajectory is described in Algorithm~\ref{alg:genTraj}. The input to the algorithm is any initial path through the racing circuit, 
parameterized in terms of distance along the path $s$, path curvature $K(s)$, and the lane edge distances $w_\mathrm{in}(s)$ and 
$w_\mathrm{out}(s)$ described in Fig.~\ref{fig:worldInfo}. Given the initial path, the minimum-time speed profile $U_x(s)$ is calculated as described in 
Fig.~\ref{fig:VPgen}. Next, the path is modified by solving the previously described minimum curvature
 convex optimization problem (\ref{eq:OPT}). 
 
 \begin{algorithm}
 \caption{Method for Iterative Trajectory Generation}
  \label{alg:genTraj}
\begin{algorithmic}[1]
\Procedure{GenerateTrajectory}{$s^\circ, K^\circ, w_\mathrm{in}^\circ, w_\mathrm{out}^\circ$}
\State $\mathrm{path}\gets (s^\circ, K^\circ, w_\mathrm{in}^\circ, w_\mathrm{out}^\circ)$
\While{$\Delta t^\star > \epsilon$}
\State $U_x \gets \mathrm{calculateSpeedProfile(path)}$
\State $\mathrm{path} \gets \mathrm{minimizeCurvature}(U_x, \mathrm{path})$
\State $t^\star \gets \mathrm{calculateLapTime}(U_x,\mathrm{path})$
\EndWhile
\State \textbf{return} $\mathrm{path},U_x$
\EndProcedure
\end{algorithmic}
\end{algorithm}
 
The optimization only solves explicitly for the steering input $\delta^\star$ and resulting vehicle lateral states $x^\star$ at every 
time step. Included within $x^\star$  is the optimal vehicle heading $\Psi^\star$ and lateral deviation $e^\star$ from the initial path. To obtain 
the new path in terms of $s$ and $K$, the East-North coordinates ($E_k$, $N_k$) of the updated vehicle path are updated as follows:
	
\begin{subequations}
\begin{align}
	E_k &\gets E_k - e^\star_k\cos(\Psi_{\mathrm{r},k})\\
	N_k &\gets N_k - e^\star_k\sin(\Psi_{\mathrm{r},k})
\end{align}
\end{subequations}
where $\Psi_\mathrm{r}$ is the path heading angle of the original path. Next, the new path is given by the following numerical approximation:
 \begin{subequations}
 \label{eq:pupdate}
\begin{align}
	s_k &= s_{k-1} + \sqrt{(E_k - E_{k-1})^2 + (N_k - N_{k-1})^2}\\ 
	K_k &= \frac{\Psi^\star_k - \Psi^\star_{k-1}}{s_k - s_{k-1}}
\end{align}
\end{subequations}
 Notice that (\ref{eq:pupdate}) accounts for the change in the total path length that occurs when the vehicle deviates from the original path.
 In addition to $s$ and $K$, the lateral distances to the track edges $w_\mathrm{in}$ and $w_\mathrm{out}$ are different for the new path
 as well, and are recomputed using the Cartesian coordinates for
the inner and outer track edges and ($E_k$, $N_k$). The two-step procedure is iterated until the improvement in lap time $\Delta t^\star$ over the prior iteration is less than a small positive 
constant $\epsilon$.

 \begin{figure}
\centering
\includegraphics[width=3.5 in]{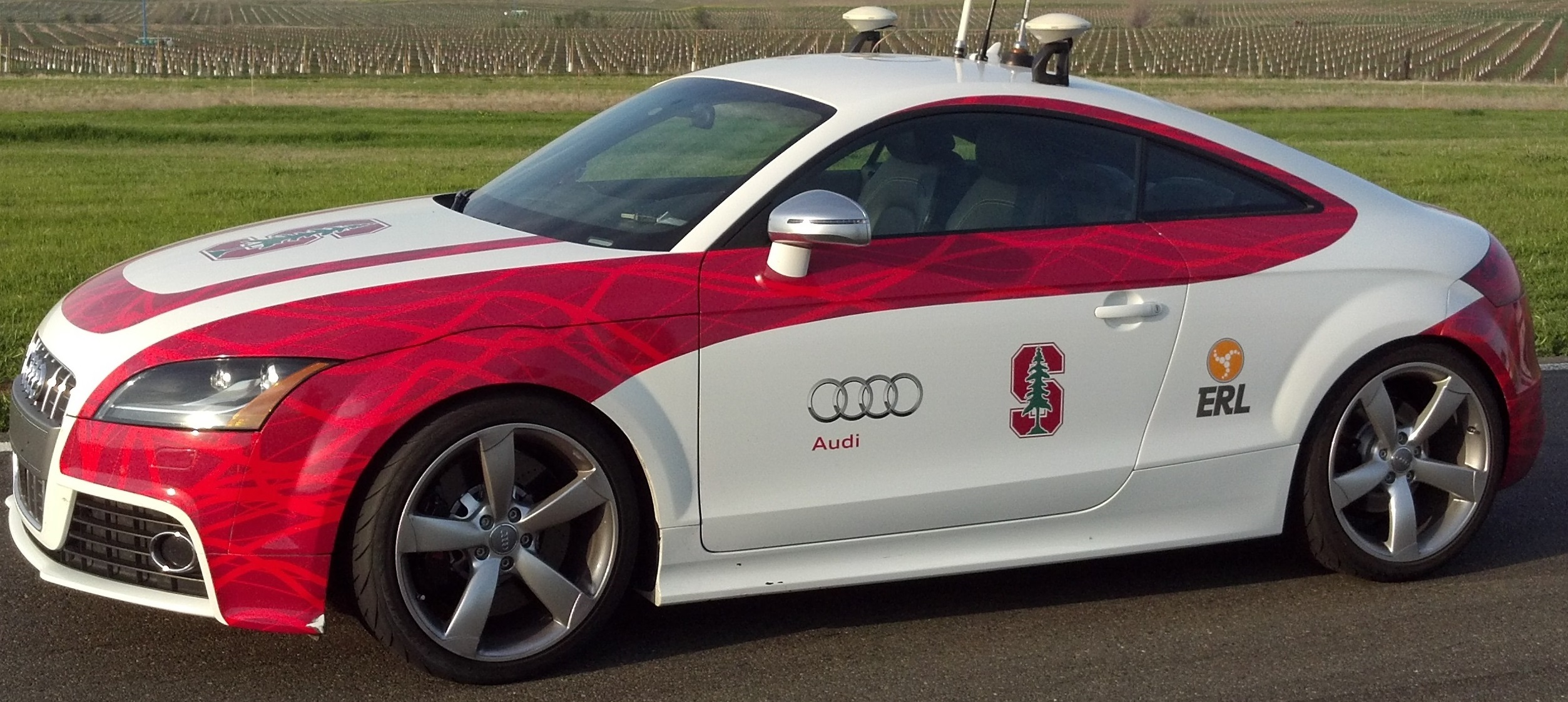}
\caption{Audi TTS used for simulation parameters and experimental validation}
\label{fig:shelleyPic}
\end{figure}

\begin{table}[h]
\begin{center}
\small
\caption{Optimization Parameters}\label{tb:params}
\begin{tabular}{lccc}
Parameter & Symbol & Value & Units \\\hline
Regularization Parameter & $\lambda$ & 1 & $\mathrm{1/m}^2$\\
Stop Criterion           & $\epsilon $ & .1 & s \\
Vehicle mass & $m$ & 1500 & kg \\
Yaw Inertia & $I_z$ & 2250 & $\mathrm{kg \cdot m}^2$\\
Front axle to CG & $a$ & 1.04 & m\\
Rear axle to CG & $b$ & 1.42 & m\\
Front cornering stiffness & $\mathrm{C}_\mathrm{f}$ & 160 & $\mathrm{kN \cdot rad}^{-1}$ \\
Rear cornering stiffness & $\mathrm{C}_\mathrm{r}$ & 180 & $\mathrm{kN \cdot rad}^{-1}$ \\
Friction Coefficient     & $\mu $                  &  0.95 & $\mathrm{-} $ \\
Path Discretization      & $\Delta s$              & 2.75 & $m$\\
Optimization Time Steps  & $T       $              & 1843 & -  \\
Max Engine Force         & -                       & 3750 & N\\\hline
\end{tabular}
\end{center}
\end{table}

\subsection{Algorithm Validation}
The proposed algorithm is tested on the 4.5 km Thunderhill racing circuit in Willows, California, USA. The vehicle parameters 
used for the lap time optimization come from an Audi TTS experimental race vehicle (Fig.~\ref{fig:shelleyPic}), and are shown along with the optimization parameters in Table 1.
 The initial path is obtained by collecting GPS data of the inner and outer track edges and estimating the ($s, K, w_\mathrm{in}, w_\mathrm{out}$)
 parametrization of the track centerline via a separate curvature estimation subproblem similar to the one proposed in \cite{perantoni}.
 The algorithm is implemented in MATLAB, with the minimum curvature optimization problem (\ref{eq:OPT}) solved using the CVX software package \cite{boydcvx}.

 \begin{figure*}
\centering
\includegraphics[width=4.5in]{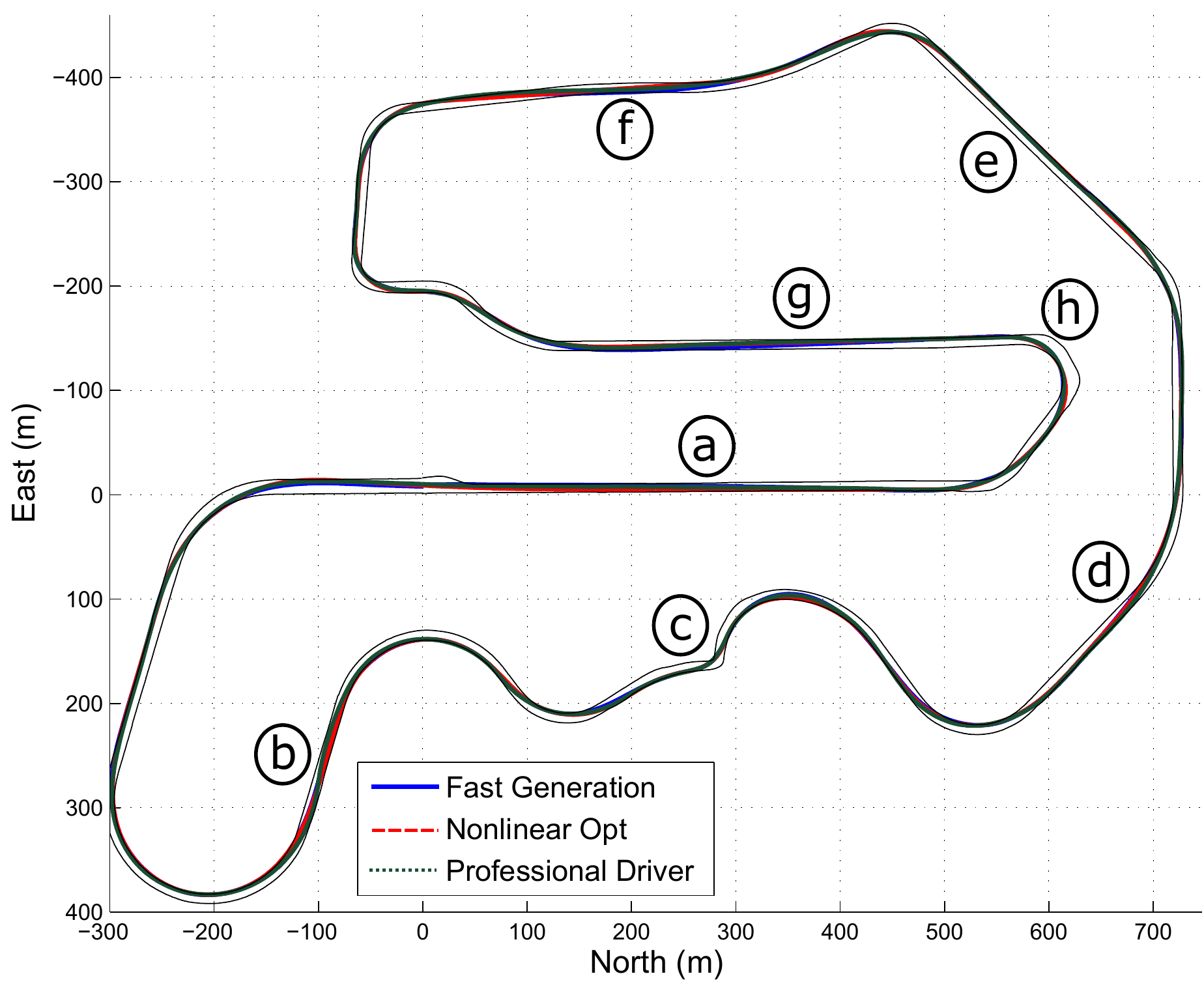}
\caption{Overhead view of Thunderhill Raceway along with generated path from algorithm. Car drives in alphabetical direction around the closed circuit. Labeled regions a-h are locations of discrepancies between the 
two-step algorithm solution and comparison solutions.}
\label{racingLines}
\end{figure*}

\begin{figure*}
\centering
\includegraphics[width=6.5in]{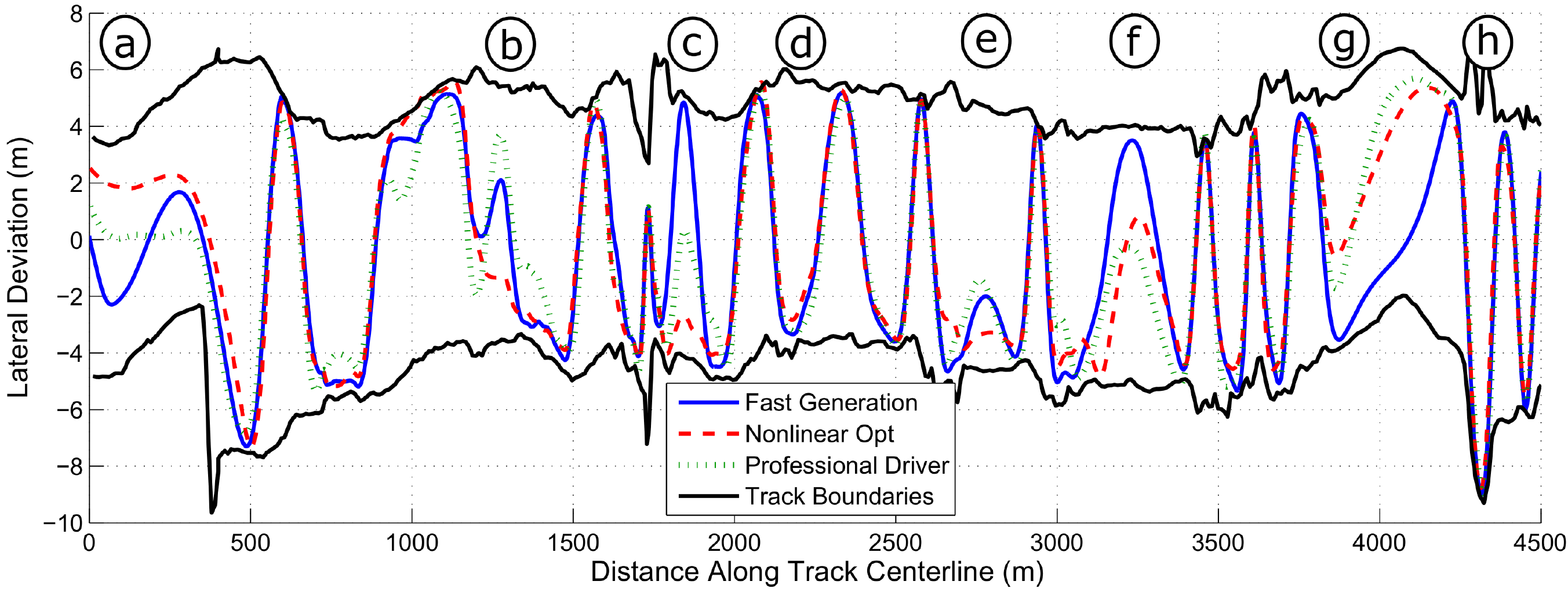}
\caption{Lateral path deviation of racing line from track centerline as a function of distance along the centerline. Note that upper and lower bounds on $e$ are not always symmetric
due to the initial centerline being a smooth approximation. Results
are compared with racing line from a nonlinear gradient descent algorithm and experimental data recorded from a professional racecar driver.}
\label{fig:pathDeviation}
\end{figure*}

\subsection{Comparison with Other Methods}
The generated racing path after five iterations is shown in
Fig.~\ref{racingLines}. To validate the proposed algorithm, the racing line is compared with results from a nonlinear gradient descent algorithm implemented by 
Theodosis and Gerdes \cite{theodosis} and an experimental trajectory recorded from a professional racecar driver in the testbed vehicle (Fig.~\ref{fig:shelleyPic}).
 While time-intensive to compute, the gradient descent approach generates racing lines with autonomously driven lap times within one second of
lap times measured from professional racecar drivers. 

To better visualize the differences between all three racing lines, Fig.~\ref{fig:pathDeviation} shows the lateral deviation from the track centerline as a function of distance along the centerline 
for all three trajectories. The left and right track boundaries $w_\mathrm{in}$ and $w_\mathrm{out}$ are plotted as well. Note that the two-step iterative algorithm provides a 
racing line that is qualitatively similar to the racing lines provided by the nonlinear gradient descent and human driver data. In particular, all three
solutions succeed at effectively utilizing all of the available track width whenever possible, and strike similar apex points on each
of the circuit's 15 corners. 

However, there are several locations on the track where there is a significant discrepancy on the order of several meters between the two-step algorithm's trajectory and the 
other comparison trajectories. These locations of interest are labeled \circled{a} through \circled{h} in Fig.~\ref{racingLines}.
Note that sections \circled{a}, \circled{e}, \circled{f}, and \circled{g} all occur on large, relatively straight portions of the racing circuit.
In these straight sections, the path curvature is relatively low and differences in lateral deviation from the track centerline have a relatively small
effect on the lap time performance.   

\begin{figure}[h]
\centering
\includegraphics[width=2.25in]{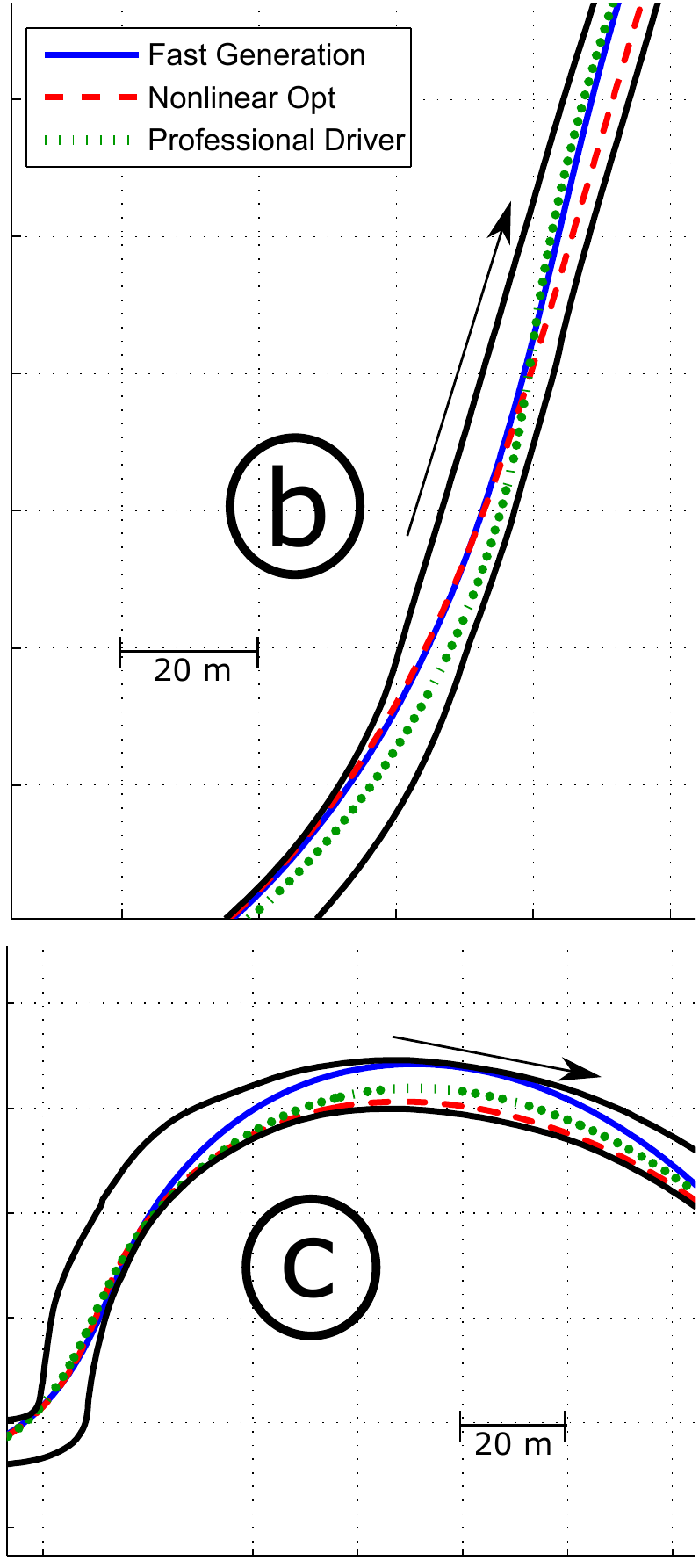}
\caption{Racing lines from the two-step fast generation approach, nonlinear gradient descent algorithm, and experimental data taken
from professional driver. Car drives in direction of labeled arrow.}
\label{compositeFig1}
\end{figure}	

\begin{figure}
\centering
\includegraphics[width=2.25in]{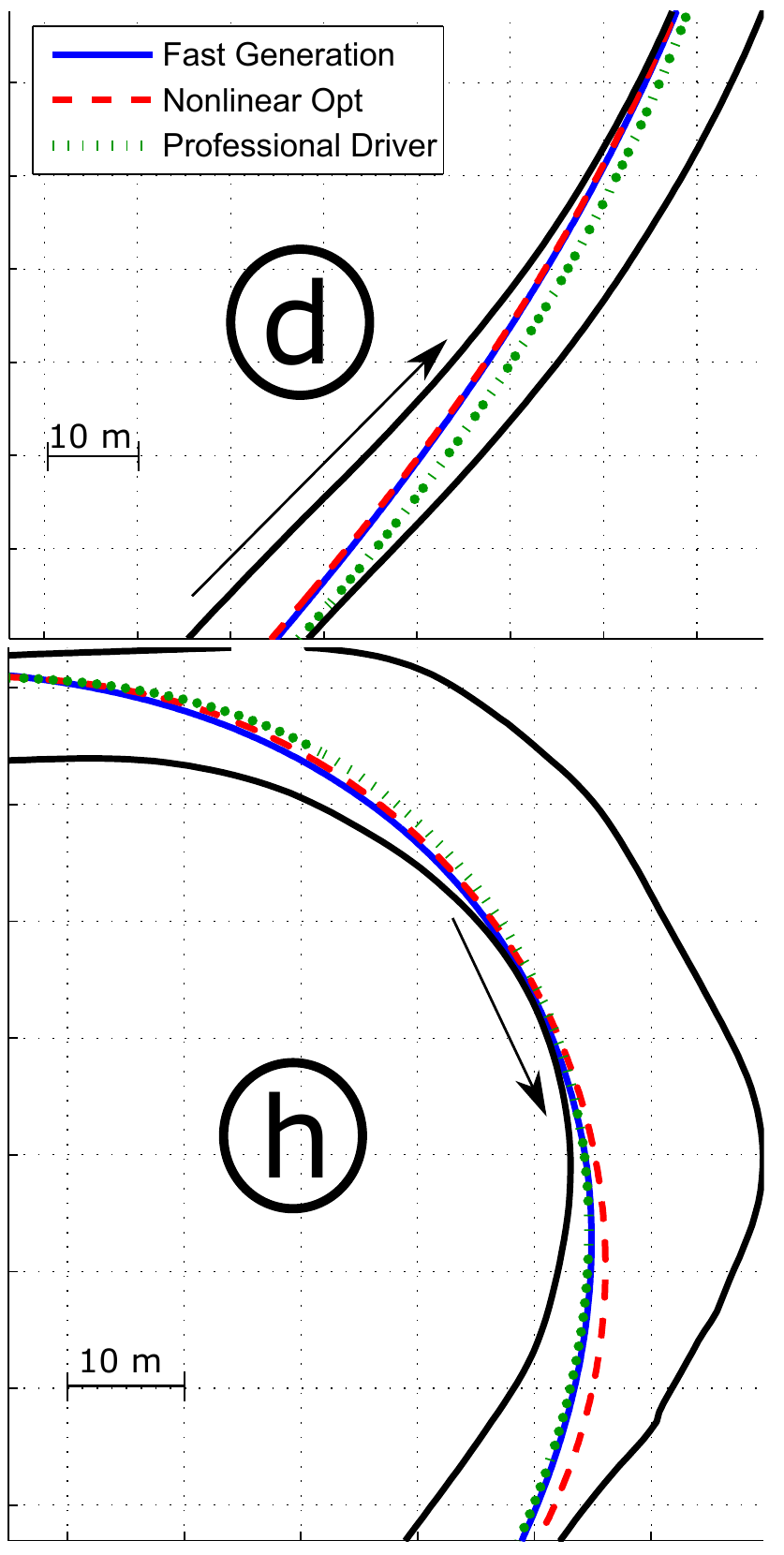}
\caption{Racing lines from the two-step fast generation approach, nonlinear gradient descent algorithm, and experimental data taken
from professional driver. Car drives in direction of labeled arrow.}
\label{compositeFig2}
\end{figure}

Of more significant interest are the sections labeled \circled{b}, \circled{c}, \circled{d}, and \circled{h}, which all occur at turning regions of the track.
These regions are plotted in Fig.~\ref{compositeFig1} and Fig.~\ref{compositeFig2} for zoomed-in portions of the race track. While it is difficult to analyze a single turn
of the track in isolation, discrepancies can arise between the two-step fast generation method and the gradient descent as the latter method trades off 
between a minimum curvature path and the path with shortest total distance. As a result, the gradient descent method finds
 regions where it may be beneficial to use less of the available road width in order to reduce the total distance
traveled. 

\begin{figure}
\centering
\includegraphics[width=3in]{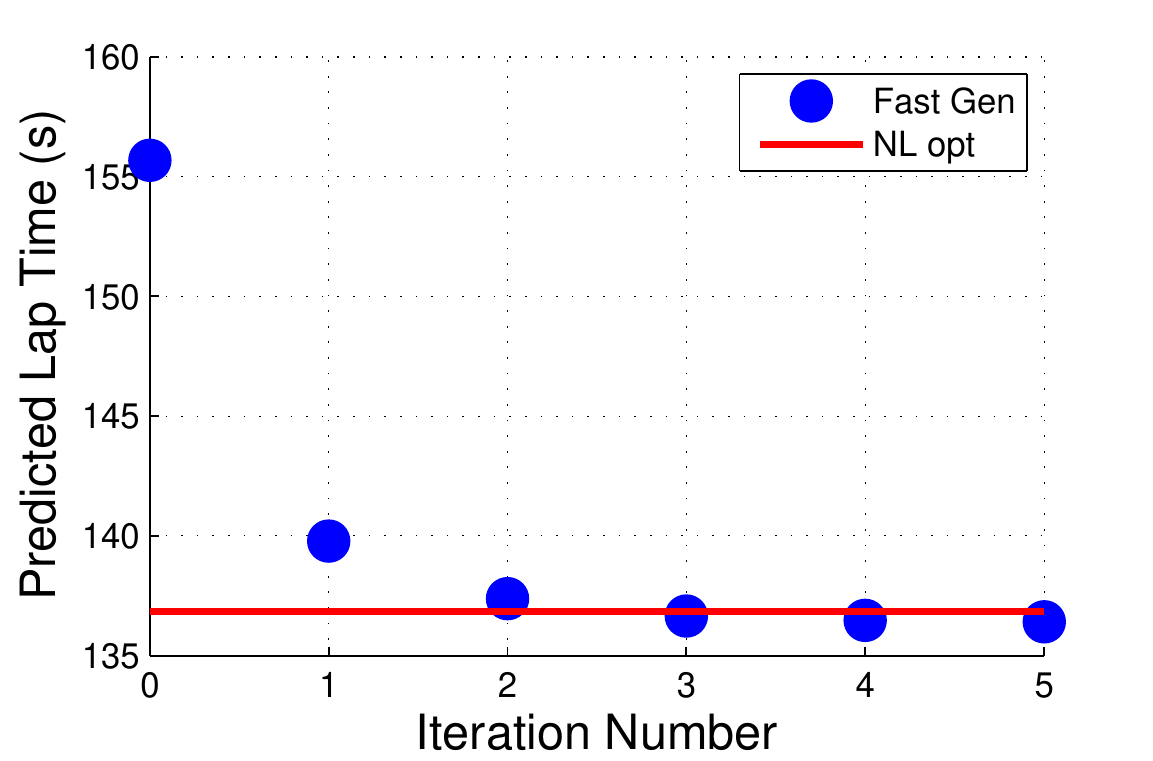}
\caption{Lap time as a function of iteration for the two-step fast trajectory generation method. Final lap time is comparable
to that achieved with the nonlinear gradient descent approach. Iteration zero corresponds to the lap time for driving the track centerline.}
\label{lapTimes}
\end{figure}

\begin{figure*}[tb]
\centering
\includegraphics[width=6.8 in]{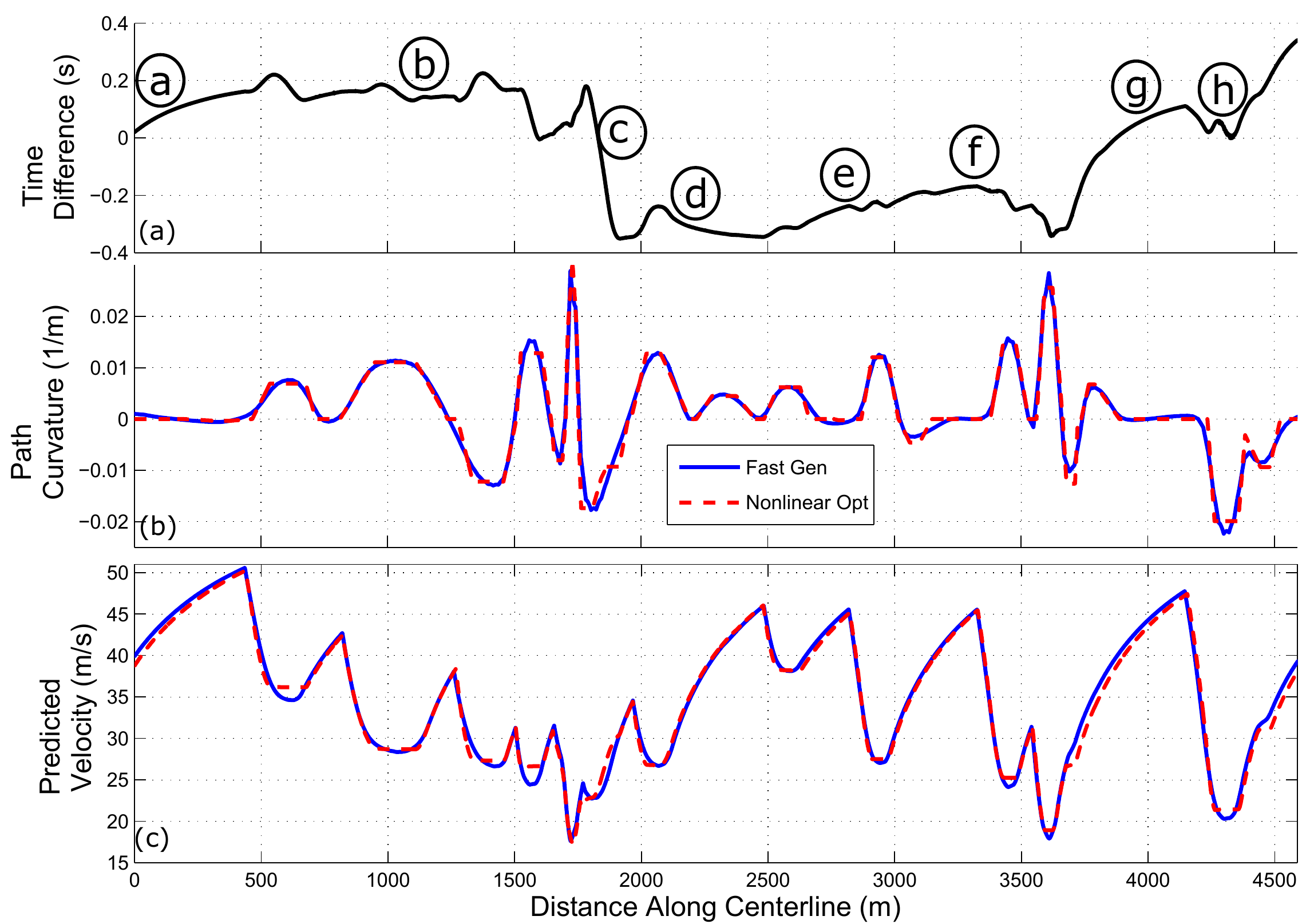}
\caption{(a) Predicted time difference between a car driving both trajectories, with a positive
value corresponding the two-step algorithm being ahead. (b) Curvature profile $K(s)$ plotted vs. distance along the path $s$.
 (c) Velocity profile $U_x(s)$ plotted 
vs. distance along the path $s$ for the two-step method and nonlinear gradient descent method.}
\label{fig:simData}
\end{figure*}

In region \circled{b}, for example, the fast generation algorithm exits the turn and gradually approaches the left side in order to create space for the
upcoming right-handed corner.  The nonlinear optimization, however, chooses a racing line that stays toward the right side of the track. In this case,
the behavior of the human driver more closely matches that of the two-step fast generation algorithm. The human driver also drives closer to the 
fast generation solution in \circled{h}, while the gradient descent algorithm picks a path that exits the corner with a larger radius. In section \circled{c}, the gradient descent
algorithm again prefers a shorter racing line that remains close the the inside edge of the track, while the two-step algorithm drives all the way
to the outside edge while making the right-handed turn. Interestingly, the human driver stays closer to the middle of the road, but more closely
follows the behavior of the gradient descent algorithm. However, there are also regions of the track where the computational algorithms pick a similar path that 
differs from the human driver, such as region \circled{d}.

\subsection{Lap Time Convergence and Predicted Lap Time}
Fig.~\ref{lapTimes} shows the predicted lap time for each iteration of the fast generation algorithm, with step 0 corresponding
to the initial race track centerline trajectory. The lap time was estimated after each iteration by numerically simulating a vehicle 
 following the desired path and velocity profile using a closed-loop controller. The equations of motion for the simulation
were the nonlinear versions of (\ref{eq:bm}) with tire forces given by
the brush tire model in (\ref{eq:fiala}). 

Fig.~\ref{lapTimes} shows that the predicted lap time converges monotonically over four or five iterations, with significant 
improvements over the centerline trajectory occuring over the first two iterations. The predicted minimum lap time of 136.4 seconds
is similar to the predicted lap time of 136.7 seconds from the nonlinear gradient descent approach, 
although in reality, the experimental lap time will depend significantly on unmodelled effects such 
as powertrain dynamics.  

The final curvature and velocity profile  for the two-step fast generation method is compared with the equivalent profiles for the gradient
descent algorithm in Fig.~\ref{fig:simData}. Notice that the piecewise linear nature of the 
nonlinear gradient descent method is due to the clothoid constraint imposed by Theodosis and Gerdes \cite{theodosis} for ease of autonomous path following.
In general, the curvature and velocity profiles are very similar, although the fast generation algorithm results in a velocity profile
with slightly lower cornering speeds but slightly higher top speeds. The predicted time difference between a car driving both trajectories 
is shown in Fig.~\ref{fig:simData}(a), with a positive
value corresponding the two-step algorithm being ahead. The trajectory from the two-step algorithm is predicted to outperform the 
gradient descent trajectory from \circled{a}-\circled{c}, lose time from \circled{c}-\circled{e}, and gain time from \circled{e}-\circled{h}.

\begin{figure*}
\centering
\includegraphics[width=6.5 in]{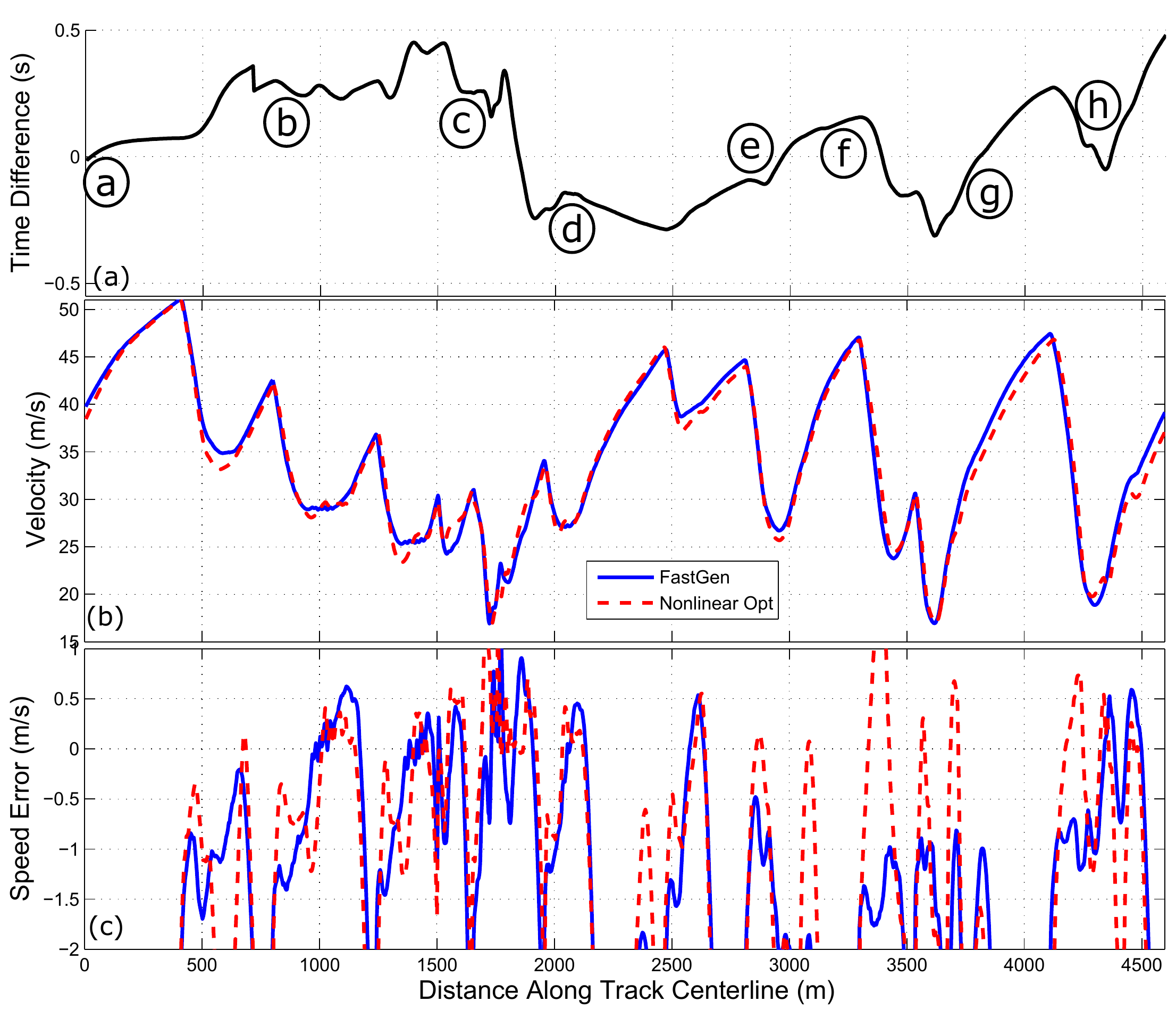}
\caption{Experimental data for an autonomous vehicle driving the trajectories provided by the two-step fast generation and gradient descent algorithms.(a) Relative time difference
between vehicle driving both trajectories, with a positive time corresponding to the two-step algorithm being ahead. (b) Actual
recorded velocity of vehicle. (c) Difference between actual and desired speed. Large negative values outside plotting range occur 
 on straight sections of the track where the vehicle is limited by engine power and speed tracking error is poorly defined.}
\label{fig:expdata}
\end{figure*}

\section{Experimental Validation} 
\label{sec:EXP}
While the two-step algorithm works well in simulation, the most critical validation step is to have an
autonomous race car drive the generated trajectory. This was accomplished by collecting experimental data on ``Shelley" (Fig.~\ref{fig:shelleyPic}), an
Audi TTS developed jointly by Stanford University and Audi's Electronics Research Laboratory (ERL). The TTS is equipped 
with an electronic power steering motor for autonomous steering and active brake booster and throttle by wire for
longitudinal control. 

Autonomous closed-loop following of the racing trajectory is accomplished by using an integrated Differential Global Positioning System (DGPS) 
and Inertial Measurement Unit (IMU) to obtain the global vehicle position and velocity. A localization algorithm is applied to find the vehicle's position along the desired path ($s$) and lateral/heading deviation ($e$ and $\Delta\Psi$) from the path. A previously developed feedback-feedforward steering algorithm \cite{kapaniaVSD} 
is then applied to keep the vehicle following the desired path at high lateral and longitudinal accelerations. By locating the desired position along the path, the current vehicle speed can be referenced to the desired vehicle speed from Fig.~\ref{fig:simData}, 
and a simple proportional speed tracking controller with feedforward can be applied to track the desired speed profile. The same 
controller setup is also used to experimentally drive the trajectory generated by the nonlinear gradient descent algorithm. 
Closed loop control is applied at a sample rate of 200 Hz using a dSPACE MicroAutobox unit.  The experimental data 
for an autonomous lap of driving is shown in Fig.~\ref{fig:expdata} for both the two-step trajectory and the trajectory from the gradient descent. 

The resulting experimental lap time for the iterative two-step algorithm was 138.6 seconds, about 0.6 seconds faster
than the experimental lap time for the gradient descent algorithm (139.2 seconds). For safety reasons, the trajectories were generated using a conservative peak road friction value of $\mu = 0.90$,
 resulting in peak lateral and longitudinal accelerations of 0.9$g$. In reality, the true friction value of the road varies slightly, but is closer to $\mu = 0.95$ on average. As a
result, both of these lap times are slightly slower than the fastest lap time recorded by a professional race car driver (137.7 seconds) and the predicted lap times from Section \ref{sec:IMPLEMENT}. A summary of all lap times is provided in Table \ref{tb:laptimes}.

\begin{table}[h]
\begin{center}
\begin{tabular}{c|cc}
    & Simulation & Experiment \\\hline
Fast Generation& 136.4 & 138.6 \\
Gradient Descent&  136.7 & 139.2 \\
Human Driver& N/A & 137.7 \\\hline
\end{tabular}
\caption{Lap Times in Seconds}\label{tb:laptimes}
\end{center}
\end{table}
The experimental data in Fig.~\ref{fig:expdata} generally matches the simulated results in Fig.~\ref{fig:simData}. The simulation predicted the trajectory from the iterative two-step algorithm would be 0.3 seconds
 quicker than that of the nonlinear algorithm, compared to the 0.6 second speed advantage observed experimentally. 
 The simulation also predicted a relative time advantage for the two-step algorithm from sections \circled{a} to \circled{c}
 and from \circled{e} to \circled{h}, a trend seen in the experimental data as well. The two-step algorithm has relatively poor performance
 from section \circled{c} to \circled{d}. This portion of the track
corresponds to the sharp right-handed turn shown in 
Fig.~\ref{compositeFig1}, where the two-step solution differs significantly from the human driver data and the gradient descent solution. This
turn also occurs on a steep downhill segment of the track, which was not accounted for by the fast generation algorithm. The experimental results
 indicate the minimum curvature heuristic is relatively poor for this particular turn when compared to a nonlinear algorithm 
 that explicitly minimizes lap time. Accounting for three-dimensional topography effects in the curvature minimization or adding a term in the cost function to minimize
 distance traveled may improve the performance in the future.
 
 Another reason for variation between the simulated and experimental time difference plots is variation in speed tracking. 
 The speed tracking error for both racing lines is shown in Fig.~\ref{fig:expdata}(c). Interestingly, while the same speed tracking controller was used to test both racing lines, 
 the controller has slightly better speed tracking performance when running the trajectory from the nonlinear optimization. This is possibly due to the
 longitudinal controller gains being originally tuned on the trajectory from Theodosis and Gerdes \cite{theodosis}.

\section{Discussion and Future Work}
\label{sec:DISCUSSION}

The primary benefit of the proposed algorithm is not improved lap time performance
over the nonlinear algorithm but rather a radical improvement in computational simplicity and speed. Each two-step iteration of the full
course takes only 26 seconds on an Intel i7 processor, whereas the nonlinear algorithm from \cite{theodosis} typically runs over the course of several hours on the same
machine. The most significant computational expense for the proposed algorithm is solving the convex curvature minimization problem for
all 1843 discrete time steps $T$ over the 4.5 km racing circuit.

\begin{table}[h]
\begin{center}
\caption{Iteration Computation Time}\label{tb:solvetime}
\begin{tabular}{ccc}
Lookahead (m) & $T$ & Solve Time (s) \\\hline
450& 184 &5 \\
900&  369 & 6 \\
1800& 737 & 12 \\
4500& 1843& 26 \\\hline
\end{tabular}
\end{center}
\end{table}
This computational efficiency will enable future work to incorporate the trajectory modification algorithm as an online
 ``preview" path planner, which would provide the desired vehicle trajectory for an upcoming portion of the race track. 
 Since the computation time of the algorithm is dependent on the preview distance, 
 the high-level planner would not need to run at the same sample time as the vehicle controller. Instead, the planner would operate on a separate CPU and provide a velocity profile and racing line for only
 the next 1-2 kilometers of the race track every few seconds, or plan a path for the next several hundred meters within a second.
 
 Table \ref{tb:solvetime} shows problem solve times for a varying range of lookahead lengths with the same discretization $\Delta s$,
 and shows that the runtime scales roughly linearly with the lookahead distance. The above solve times are listed using the CVX convex optimization solver, which 
 is designed for ease of use and is not optimized for embedded computing. Preliminary work has been successful
 in implementing the iterative two-step algorithm into C code using the CVXGEN software tool \cite{boydcvxgen}. When written
 in optimized C code, the algorithm can solve the curvature minimization problem (\ref{eq:OPT}) in less than 0.005 seconds for a lookahead distance of 650 meters.
 
 The possibility of real-time trajectory planning for race vehicles creates several fascinating areas of future research. An automobile's surroundings are subject to both rapid and
 gradual changes over time, and adapting to unpredictable events requires a real-time trajectory planning algorithm.
 On a short time scale, the real-time trajectory planner could find a fast but stable recovery trajectory in the event of the race vehicle entering an understeer or oversteer 
 situation. On an intermediate time scale, the fast executing two-step algorithm could continuously plan a racing line in the presence of other moving race vehicles by constraining the
 permissible driving areas to be collision-free convex ``tubes" \cite{erlien}.  Finally, the algorithm could update the racing trajectory given estimates of the friction coefficient and other vehicle parameters learned gradually over
 several laps of racing.
 
\section{Conclusion}
This paper demonstrates an iterative algorithm for quickly generating vehicle racing trajectories, where each iteration is comprised of
 a sequential velocity update and path update step. Given an initial path through the race track, the 
 velocity update step performs forward-backward integration to determine the minimum-time speed inputs. Holding this speed 
 profile constant, the path geometry is updated by solving a convex optimization problem to minimize path curvature. Experimental data confirms that the results
 generated by the algorithm for the Thunderhill Raceway circuit are comparable to those from a nonlinear gradient descent algorithm, with the primary advantage being a much
 faster computation time. An exciting opportunity for future research is incorporating the trajectory modification algorithm into an online path planner to provide
 racing trajectories in real time.

\begin{acknowledgment}
The authors would like to thank Marcial Hernandez for assistance with fitting an initial curvature profile from GPS point
cloud data, and Vadim Butakov and Rob Simpson from the Audi Electronics Research Lab in Belmont, CA. The authors would
also like to thank Vincent Laurense, Samuel Schacher, and Jon Pedersen for assistance with autonomous data collection.
Kapania and Subosits are both supported by Stanford Graduate Fellowships. 
\end{acknowledgment}

\bibliographystyle{asmems4}
\bibliography{jdsmc2015}

\end{document}